\documentclass{article}

\usepackage{PRIMEarxiv}

\usepackage[utf8]{inputenc} 
\usepackage[T1]{fontenc}    
\usepackage{hyperref}       
\usepackage{url}            
\usepackage{booktabs}       
\usepackage{amsfonts}       
\usepackage{nicefrac}       
\usepackage{microtype}      
\usepackage{lipsum}
\usepackage{fancyhdr}       
\usepackage{graphicx}       
\graphicspath{{media/}}     

\usepackage{amsmath}
\usepackage{appendix}

\title{ZLPR: A Novel Loss for Multi-label Classification
}

\author{
  Jianlin Su, Mingren Zhu, Ahmed Murtadha, Shengfeng Pan, Bo Wen, Yunfeng Liu \\
  \\
  Zhuiyi Technology Co., Ltd. Shenzhen, China \\
  \\
  \texttt{\{bojonesu, zhumingren, mengjiayi, nickpan, brucewen, glenliu\}@wezhuiyi.com} \\
}

\begin{document}
\maketitle

\begin{abstract}
In the era of deep learning, loss functions determine the range of tasks available to models and algorithms.
To support the application of deep learning in multi-label classification (MLC) tasks, we propose the ZLPR (zero-bounded log-sum-exp \& pairwise rank-based) loss in this paper.
Compared to other rank-based losses for MLC, ZLPR can handel problems that the number of target labels is uncertain, which, in this point of view, makes it equally capable with the other two strategies often used in MLC, namely the binary relevance (BR) and the label powerset (LP).
Additionally, ZLPR takes the corelation between labels into consideration, which makes it more comprehensive than the BR methods.
In terms of computational complexity, ZLPR can compete with the BR methods because its prediction is also label-independent, which makes it take less time and memory than the LP methods.
Our experiments demonstrate the effectiveness of ZLPR on multiple benchmark datasets and multiple evaluation metrics.
Moreover, we propose the soft version and the corresponding KL-divergency calculation method of ZLPR, which makes it possible to apply some regularization tricks such as label smoothing to enhance the generalization of models.
\end{abstract}


\section{Introduction}\label{s1}

Multi-label classification (MLC) tasks are very popular in data science for now and it has a very wide range of applications such as multi-object detection for images\cite{2017Faster}, multi-topic discovery for texts\cite{2020Federated}, multi-function classification for genes\cite{2021AC}, and so on\cite{Christel2022Multilabel}.
The framework of MLC can be expressed as following:
\begin{itemize}
  \item it includes a sample space $\mathcal{X}$ and a label space $\mathcal{Y}$, where the label space is a combination of a base category set $\Lambda = \{\lambda_l\}_{l = 1}^L$; 
  \item it includes a dataset $D_{train} = (X, Y)$ for training with both the samples $X = \{x_n| x_n \in \mathcal{X}\}_{n = 1}^N$ and the labels $Y = \{y_n| y_n \in \mathcal{Y}\}_{n = 1}^N$; 
  \item it includes another dataset $D_{test} = (X', )$ to predict with only the samples $X' = \{x_m'| x_m' \in \mathcal{X}\}_{m = 1}^M$; 
  \item what we are going to do is training a model $f_{\theta}: \mathcal{X} \rightarrow \mathcal{Y}$ with $D_{train}$ and use it to get appropriate labels $Y' = \{\hat{y}_m'| \hat{y}_m' = f_{\theta}(x_m'), x_m' \in X'\}_{m = 1}^M$ for $D_{test}$.
\end{itemize}
This is not really different from "normal" classification tasks, which is usually called the multi-class classification (MCC)\footnote{
  The binary classification is regarded as a special case of MCC ($L = 2$).
}.
However, the label space of an MCC task has strong mutual exclusion, which makes the labels $y^{mcc}$ all have and only have one base category, i.e., $|y^{mcc}| \equiv 1$.
On the contrary, MLC tasks allow one or more category elements to appear at the same time, or no elements at all.
The label space of an MCC task is just its own base label set, i.e., $\mathcal{Y}^{mcc} = \Lambda^{mcc}$, while the label space of an MLC task can be the power set of its base category set, i.e., $\mathcal{Y}^{mlc} = 2^{\Lambda^{mlc}}$.
It can be seen that single-label MCC is a special case of the MLC framework.

There are several challenges faced by research in MLC.
In addition to the category imbalance problem\cite{2017AnB} commonly found in classification, MLC tasks are also disrupted by other three issues due to the nature of their label space.
First of all, the label of a data sample in MLC is actually a set of basic category elements, which, analogous to the one-hot vector in MCC, can be represented by a "multi-hot" vector.
It would be simpler if the number of 1s in the multi-hot vector was fixed, but for most MLC tasks, the size of labels is indeterminate.
For example, in the same MLC task, the labels of some data samples may contain multiple category elements, some may have only one, and some may not even have one.
This adds difficulty to the modeling of the task.

Second, the plurality of data labels may imply correlation information between categories.
For example, in an image object detection task, cats and dogs may often appear together, whereas cats and sharks do not.
This is in stark contrast to MCC, where there is a known clear mutual exclusivity between categories in MCC tasks but correlations between categories in MLC tasks require further modeling or learning from data.
Modeling and learning between category correlations has always been one of the core research problems of the MLC community\cite{2020Fast}.

The last thing to point out is the sparsity of labels in MLC, which is actually a further result of the previous two problems.
Since the number of categories of samples is uncertain, the size of the label space in MLC is potentially exponential to the size of the base category set, i.e., $|\mathcal{Y}^{mlc}| \sim 2^{|\Lambda^{mlc}|}$.
However, due to unknown correlations between categories, some labels in the exponential label space may never appear, while others appear frequently.
This leads to the sparsity of the label space and the imbalance between labels\cite{2014Towards}, which, to a certain extent, replaces the problem of category imbalance.

In the traditional machine learning community, various methods have been proposed for the above issues in MLC\cite{bogatinovski2022comprehensive}.
These methods are classified into two categories: algorithm adaption (AD) methods and problem transformation (PT) methods.
AD methods primarily focus on expanding the MCC algorithms to address MLC tasks, such as MLKNN\cite{2021Multi}, MMP\cite{4634206} and Rank-SVM\cite{2008Calibrated}.
PT methods transforms MLC tasks into one or more single-label classification\cite{2016DeepBE}, or label ranking tasks\cite{2015Deep}.
In terms of flexibility, AD methods seem to be more rigid because each adaptive algorithm can only work on a specific model, while PT methods allow various models to be used for handling the task after transformation.

With the widespread success of deep learning, more and more MLC tasks are starting to work using end-to-end neural networks as base models, which is also the result of the structure of neural networks that are naturally suitable for MLC tasks\cite{2014Deep}.
When applying deep learning to deal with MLC tasks, in addition to using the structure advantages of the neural network itself, it is often combined with PT methods to enhance the processing power of the model.
Such a combination is often reflected in the design of the loss function, because it is the most task-relevant part in the deep learning architecture.
However, the loss functions proposed in previous studies either only meet partial requirements\cite{2017Focal} or go through complex multi-stage processing\cite{2017Improving}, lacking flexibility and effectiveness to deal with the above three problems in MLC.
To fill such a research gap and better support the application of deep learning in MLC tasks, we propose the ZLPR (Zero-bounded Log-sum-exp \& Pairwise Rank-based) loss in this paper:
\begin{equation}\label{eq1}
  \mathcal{L}_{zlpr} = \log\left(1+\sum_{i \in \Omega_{pos}}e^{-s_i}\right) + \log\left(1+\sum_{j \in \Omega_{neg}}e^{s_j}\right), 
\end{equation}
where $\Omega_{pos}$ is the label (set) and $\Omega_{neg} = \Lambda/\Omega_{pos}$, and $s_i$ is the model output score of the $i$-th category ($\lambda_i$).
During prediction, $s_i > 0$ implies that $\lambda_i$ could be a target category and $s_i < 0$ doesn't, which is the meaning of "Zero-bounded".

More details about ZLPR and how it handles the above three common challenges in MLC tasks will be explained in Sec.\ref{s3}.
Before that, we will review some loss functions commonly used in deep learning for MLC tasks in Sec.\ref{s2}.
In Sec.\ref{s4}, we will compare the effects of these losses on multiple datasets with multiple evaluation metrics.
Sec.\ref{s5} is the conclusion of this paper.
Additionally, we propose the soft version and the corresponding calculation method of KL-divergence of ZLPR in the appendix.
With these two, some regularization tricks such as label smoothing\cite{2021Delving} or R-drop\cite{2021arXiv210614448L} can be applied to enhance models' generalization ability.

\section{Related Work}\label{s2}

As mentioned earlier, the purpose of this paper is to promote the application of deep learning in MLC tasks.
The traditional MLC processing methods that can be combined with neural networks are mainly problem transformation (PT) methods.
Therefore, we mainly review the work about combination of PT methods and neural network in this section.

There are four modes of PT methods, namely the Binary Relevance (BR), the Label Powerset (LP), the Classifier Chian (CC) and the Label Ranking (LR)\cite{2012article}.
The BR mode converts an MLC task into multiple independent binary classification problems, where each binary classifier determines whether the corresponding category is included in the label set or not.
When combining the BR mode with a neural network, only a binary classification loss needs to be applied to each output node.
The BR mode can handle cases where the number of target categories is uncertain based on linear complexity, and due to the simplicity, it is the most frequently mode used in MLC tasks with deep learning.
Works based on the BR mode are mainly to propose binary loss functions with various characteristics.
The baseline is the (weighted) binary cross entropy loss.
Beyond that, Lin et al.\cite{2017Focal} proposed the focal loss making the model focus on hard-to-learn samples during training, which has been applied in many fields and received lots of favorable comments.
Millerari et al.\cite{2016V} proposed the dice loss trying to optimize the F1-measurement rather than the accuracy.
Based on that, Li et al.\cite{li2019dice} proposed the self-adjusting dice loss, which, however, did not gets much credit as the original paper implied.
Menon et al.\cite{menon2020long} proposed a novel logit adjustment method to mitigate the impact of data imbalance by introducing the prior of categories.
The simplicity and ease of use of the BR mode has attracted much attention, but it assumes conditional independence between categories and can not capture the correlation within the label set.

The LP mode directly takes the subsets of the base category set in MLC tasks as the prediction output, which means there should be a total of $2^{|\Lambda|}$ output nodes of the neural network.
In this way, an MLC task is transformed into an MCC task and any MCC loss function can be used.
Although the LP mode can adaptively determine the number of target categories and fully capture the correlation within the label set, the potential exponential space complexity is unacceptable, and it also faces serious sparsity and label set imbalance problems.
Therefore, this mode is not used as much as the others.
The CC mode transforms an MLC task into a sequence prediction problem with a pre-defined category order.
When predicting whether a category belongs to the target label set or not, the CC mode use the previously obtained category prediction result as input, which forms a chain architecture to capture the correlation between categories.
The CC mode is not implemented by modifying the loss function, but by leveraging the model architecture, which can apply the RNN\cite{2016CNN} or the Seq2Seq\cite{2021History} modules in deep learning.
It should be noted that the outputs of the CC mode are in series while others are parallel, so the CC mode is not as time efficient as the BR mode and LP mode.

The LR mode is different from the above three modes in that it no longer treats MLC tasks as classification but as ranking.
The basic idea of the LR mode is requiring that the rank scores of target categories are greater than non-target categories.
Based on the framework of learning to rank\cite{2020Learning} via pairwise comparisons, some rank-based loss functions have been proposed to deal with MLC tasks in the LR mode.
Weston et al.\cite{1970Weighted} proposed the WARP loss that puts different weights on violations where the score of target category is less than non-target category.
The intuition is that if the positive category is ranked lower, the violation should be penalized higher.
Zhang et al.\cite{2008Improved} proposed the BP-MLL loss, which is essentially an exponential pairwise ranking loss.
Li et al.\cite{2017Improving} proposed the LSEP loss based on the pairwise ranking loss and proved it has favorable theoretical guarantees compared to the hinge alternative.
Sun et al.\cite{2020Circle} proposed a unified loss function for pairwise optimization, which is a generalization of the LSEP loss.
The LR mode has parallel outputs as the BR mode, square space complexity during training and linear during prediction.
Better than the BR mode, the LR mode does not introduce additional independence assumptions, thus preserving the category correlations implied by the original data during training.

Models trained by the loss functions mentioned above in the LR mode cannot directly output the target label set when making predictions.
To deal with that, there are two empirical methods often used.
The first is to manually set a threshold, and categories with scores greater than this threshold will be considered as the target label set.
The second is to directly output categories with top-k scores as the target label set.
These two methods cannot handle the problem of manual intervention or uncertainty about the number of target categories.
Li et al.\cite{2017Improving} proposed a two-stage approach to deal with the above situation, first using the LSEP loss to optimize the class scores, and then adding new modules to estimate the number of target categories or the threshold.
This two-stage approach increases the model complexity and the training instability.
The ZLPR loss proposed in this paper combines the advantages of the LR mode and the BR mode, achieves a high level of time and space efficiency, retains the information of the category correlation implied in the original data, and can adapt to the situation of uncertain number of categories.

\section{The ZLPR Loss}\label{s3}

The commonly used loss function for single-label MCC tasks is the cross entropy loss.
If logits are used to represent it, the formula is as follows:
\begin{equation}\label{eq2}
  \mathcal{L}_{ce} = -\log\frac{e^{s_i}}{\sum_{j = 1}^Le^{s_j}} = \log\left(1+\sum_{j = 1, j \neq i}^Le^{s_j - s_i}\right), 
\end{equation}
where the target category is $\lambda_i$ and $s_j$ is the logit for the $j$-th category.
Since the log-sum-exp operator is a smooth approximation of the maximum operator, i.e., 
\begin{equation}\label{eq3}
  \log\left(1+\sum_{j = 1, j \neq i}^Le^{s_j-s_i}\right) \approx \max\left(0, \underbrace{s_j-s_i, \cdots}_{\forall j \neq i}\right), 
\end{equation}
minimizing the cross entropy loss actually means to make the logits of negative categories smaller than the positive one.
We extend this point to MLC tasks and get:
\begin{equation}\label{eq4}
  \log\left(1+\sum_{j \in \Omega_{neg}}e^{s_j}\sum_{i \in \Omega_{pos}}e^{-s_i}\right) \approx \max\left(0, \underbrace{s_j - s_i, \cdots}_{i \in \Omega_{pos}, j \in \Omega_{neg}}\right), 
\end{equation}
where the formula on the left is actually the LSEP loss proposed by Li et al.\cite{2017Improving} and we derived it from a different perspective.

As mentioned in Sec.\ref{s2}, the LSEP loss lacks the ability to adapt to the variable number of target categories like other general pairwise rank-based losses.
To handle such a situation, we introduce a threshold $s_0$ in the loss, hoping that the logits of positive categories are greater than it and the negative categories is less.
In this way, we get the threshold-bounded log-sum-exp \& pairwise rank-based (TLPR) loss as follows:
\begin{equation}\label{eq5}
  \mathcal{L}_{tlpr} = \log\left(1+\sum_{j\in\Omega_{neg}}e^{s_j}\sum_{i\in\Omega_{pos}}e^{-s_i}+\sum_{j\in\Omega_{neg}}e^{s_j-s_0}+\sum_{i\in\Omega_{pos}}e^{s_0-s_i}\right), 
\end{equation}
which can also be simplified as:
\begin{equation}\label{eq6}
  \mathcal{L}_{tlpr} = \log\left(e^{-s_0}+\sum_{i \in \Omega_{pos}}e^{-s_i}\right) + \log\left(e^{s_0}+\sum_{j \in \Omega_{neg}}e^{s_j}\right).
\end{equation}
For sake of simplicity, we set $s_0 = 0$ here and get the ZLPR loss expressed in Equation \ref{eq1}.

The form of ZLPR seems to calculate the positive and negative categories separately, which does not show the characteristics of "pairwise ranking".
However, through the log-sum-exp and maximum approximation, we have:
\begin{equation}\label{eq7}
  \mathcal{L}_{zlpr} \approx \max\left(0, \underbrace{-s_i, \cdots}_{i \in \Omega_{pos}}\right) + \max\left(0, \underbrace{s_j, \cdots}_{j \in \Omega_{neg}}\right)
  = \max\left(0, \underbrace{s_j, \cdots}_{j \in \Omega_{neg}}\right) - \min\left(0, \underbrace{s_i, \cdots}_{i \in \Omega_{pos}}\right), 
\end{equation}
from which we can still get the inspiration of intuitive "pairwise ranking".

The ZLPR loss can be rewritten as:
\begin{equation}\label{eq8}
  \mathcal{L}_{zlpr} = \log\left(1+\langle y, e^{-s}\rangle\right) + \log\left(1+\langle 1-y, e^{s}\rangle\right), 
\end{equation}
where $y$ is the multi-hot label, $s$ is the logit vector and $\langle\cdot, \cdot\rangle$ is the inner product operator.

\subsection{Label Dependence}\label{s3-1}

The goal of classification algorithms in general is to capture dependencies between input features $x$ and the target label $y$, i.e., to find the conditional probability $\mathcal{P}(y|x)$.
In MLC, dependencies may not only exist between the features $x$ and the target categories $y$, but also between the categories themselves.
The idea to improve predictive accuracy by capturing such dependencies is a driving force in research on MLC\cite{2020AF}.
In this subsection, we theoretically demonstrate that the ZLPR loss has the ability to capture label dependencies from original data.

The key to this problem is determining whether it is necessary to rely on information from the joint distribution when minimizing the corresponding empirical risk of the loss function, and if the answer is yes, it means that this loss can capture label dependencies\cite{2012OnL}.
The empirical risk of ZLPR is:
\begin{equation}\label{eq9}
  \mathcal{R}_{zlpr} = \mathbb{E}_{y \sim \mathcal{P}(y|x)}\left[\log\left(1+\langle y, e^{-s}\rangle\right) + \log\left(1+\langle 1-y, e^{s}\rangle\right)\right], 
\end{equation}
and the gradient is:
\begin{equation}\label{eq10}
  \frac{\partial\mathcal{R}_{zlpr}}{\partial s} = \mathbb{E}_{y \sim \mathcal{P}(y|x)}\left[-\frac{y\odot e^{-s}}{1+\langle y, e^{-s}\rangle} + \frac{(1-y)\odot e^s}{1+\langle 1-y, e^s\rangle}\right], 
\end{equation}
where $\odot$ is the Hadamard product operator.
Setting the gradient to be zero, $\forall t \in \{1, \cdots, L\}$ we have:
\begin{equation}\label{eq11}
  s_t = \frac{1}{2}\underbrace{\log\frac{\mathcal{P}(y^{(t)} = 1|x)}{\mathcal{P}(y^{(t)} = 0|x)}}_{T_1} + \frac{1}{2}\underbrace{\log\frac{\mathbb{E}_{\tilde{y}^{(t)}\sim\mathcal{P}(\tilde{y}^{(t)}|y^{(t)} = 1, x)}\left[\phi_1(s, \tilde{y}^{(t)})\right]}{\mathbb{E}_{\tilde{y}^{(t)}\sim\mathcal{P}(\tilde{y}^{(t)}|y^{(t)} = 0, x)}\left[\phi_0(s, \tilde{y}^{(t)})\right]}}_{T_2}, 
\end{equation}
where $y^{(t)} \in \{0, 1\}$ is the binary state variable of the category $\lambda_t$, $\tilde{y}^{(t)} = \{y^{(1)}, \cdots, y^{(t-1)}, y^{(t+1)}, \cdots, y^{(L)}\}$ is the state variable set of other categories; 
$\phi_1(s, \tilde{y}^{(t)}) = [1+e^{-s_t}+\sum_{r\neq t}y^{(r)}e^{-s_r}]^{-1}$ and $\phi_0(s, \tilde{y}^{(t)}) = [1+e^{s_t}+\sum_{r\neq t}(1-y^{(r)})e^{s_r}]^{-1}$.
On the right side of the equation, $T_1$ is the margin logit of the corresponding category, and $T_2$ is the dependence coupling term.
Due to the complexity of $T_2$, we are unlikely to find a solution of $s$ such that $T_2$ does not depend on $\mathcal{P}(\tilde{y}^{(t)}|y^{(t)}, x)$, which means that $s$ will depend on the joint distribution with minimal empirical loss.
This demonstrates the ability of ZLPR to capture label dependencies.

\subsection{Comparison to Related Loss Functions}

We compare two other loss functions with ZLPR, namely the LSEP loss and the BCE loss.

\textbf{LSEP.}
The LSEP loss is expressed as:
\begin{equation}\label{eq12}
  \mathcal{L}_{lsep} = \log\left(1+\sum_{j \in \Omega_{neg}}e^{s_j}\sum_{i \in \Omega_{pos}}e^{-s_i}\right).
\end{equation}
Compared to ZLPR/TLPR, it lacks two summations for comparison with thresholds in the $\log$ operator, which makes it lose the ability to adaptively adjust the number of predicted categories.
Additionally, to make the LSEP loss scale linearly to the category size, Li et al.\cite{2017Improving} adapt the negative sampling technique\cite{2013Distributed} and sample at most $t$ pairs from the Cartesian product, which can be denoted as:
\begin{equation}\label{eq13}
  \mathcal{L}_{lsep}^{sampled} = \log\left(1+\sum_{(i, j) \in \phi(y, t)}\exp(s_j-s_i)\right), 
\end{equation}
where $\phi(y, t) \subseteq y\otimes(\Lambda/y)$ is the sampled pair set.
However, our goal is to generalize the application of deep learning in MLC, compared to the size of the current popular large models\cite{2021GPT}, the computational cost of the rank-based loss is negligible.
Therefore, we will not use the sampling method, which can also guarantee the accuracy of the results.
Finally, it should be noted that the LSEP loss can also capture label dependencies implied in the original data.

\textbf{BCE.}
The binary cross entropy (BCE) loss can be applied in MLC tasks by combining with the BR mode, which forms:
\begin{equation}\label{eq14}
  \mathcal{L}_{bce} = \sum_{i \in \Omega_{pos}}\log(1+e^{-s_i}) + \sum_{j \in \Omega_{neg}}\log(1+e^{s_j}).
\end{equation}
In fact, for the BR mode, the loss is the sum of binary losses for each category, i.e., 
\begin{equation}\label{eq15}
  \mathcal{L}_{BR} = \sum_{l = 1}^L\mathcal{L}_{b}(y^{(l)}, s_l), 
\end{equation}
and the corresponding empirical risk is 
\begin{equation}\label{eq16}
  \mathcal{R}_{BR} = \mathbb{E}_{y \sim \mathcal{P}(y|x)}\left[\mathcal{L}_{BR}\right] = \sum_{l = 1}^L\left[\mathcal{P}(y^{(l)} = 1|x)\mathcal{L}_{b}(1, s_l)+\mathcal{P}(y^{(l)} = 0|x)\mathcal{L}_{b}(0, s_l)\right], 
\end{equation}
from which we can see that the categories are not related to each other.
This is why it is said that the BR mode introduces conditional independence and fails to capture label dependencies.
Specifically for the BCE loss, the condition of minimizing the empirical risk is
\begin{equation}\label{eq17}
  \forall l = 1, \cdots, L: s_l = \log\frac{\mathcal{P}(y^{(l)} = 1|x)}{\mathcal{P}(y^{(l)} = 0|x)}.
\end{equation}

In addition, the BCE loss can be rewritten as:
\begin{align}\label{eq18}
  \mathcal{L}_{bce} &= \log\left[\prod_{i \in \Omega_{pos}}(1+e^{-s_i})\right] + \log\left[\prod_{j \in \Omega_{neg}}(1+e^{s_j})\right] \\ \nonumber
  &= \log\left[1+\sum_{i \in \Omega_{pos}}e^{-s_i}+\underbrace{\sum_{k, t \in \Omega_{pos}, k < t}e^{-(s_k+s_t)}+\cdots+e^{-\sum_{r \in \Omega_{pos}}s_r}}_{\text{positive higher-order terms}}\right] \\ \nonumber
  &~ + \log\left[1+\sum_{j \in \Omega_{neg}}e^{s_j}+\underbrace{\sum_{k, t \in \Omega_{neg}, k < t}e^{s_k+s_t}+\cdots+e^{\sum_{r \in \Omega_{neg}}s_r}}_{\text{negative higher-order terms}}\right], 
\end{align}
which, compared with the ZLPR loss in Equation \ref{eq1}, has more higher-order terms.
In MLC tasks, negative categories tends to far outnumber the positives, and these higher-order terms in BCE significantly strengthen this category imbalance problem.
Conversely, the ZLPR loss, which removes these higher-order terms, balances the influence between positive and negative categories, making training fairer across them.

\section{Experiments and Discussion}\label{s4}

\subsection{Datasets}\label{s4-1}

We mainly conduct experiments on MLC of texts, including 3 Chinese datasets and 8 English datasets:
\begin{itemize}
  \item \textbf{CNIPA\footnote{http://patdata1.cnipa.gov.cn/}:} 
    The chinese patent dataset, which has a classification system called IPC (international patent classification). 
    Each innovation patent is assigned one or more IPC codes, which can be used as the label set.
    We downloaded about 210k invention patents, and constructed two Chinese text MLC datasets using titles and abstracts, respectively called \textbf{CNIPA-Title} and \textbf{CNIPA-Abstract}.
  \item \textbf{Toutiao\footnote{https://github.com/aceimnorstuvwxz/toutiao-multilevel-text-classfication-dataset}:} 
    The chinese news dataset from Toutiao, which asks for MLC of news based on headlines.
  \item \textbf{USPTO\footnote{https://bulkdata.uspto.gov/}:} 
    The english patent dataset, which also uses the IPC system as labels.
    We downloaded about 350k invention patents, and also constructed two english text MLC datasets with titles as abstracts, respectively called \textbf{USPTO-Title} and \textbf{USPTO-Abstract}.
  \item \textbf{Archive\footnote{https://www.kaggle.com/datasets/shivanandmn/multilabel-classification-dataset}:} 
    This dataset contains 6 different categories (Computer Science, Physics, Mathematics, Statistics, Quantitative Biology, Quantitative Finance) to classify the research papers.
    We constructed two datasets based on titles and abstracts, namely \textbf{Archive-Title} and \textbf{Archive-Abstract}.
  \item \textbf{CMU-Movie\footnote{https://www.cs.cmu.edu/~ark/personas/}:} 
    This dataset contains texts about 42,306 movies extracted from Wikipedia and Freebase.
    However, it is a noisy dataset because the same movie category may have different descriptions, like "actions" and "action movie".
    We constructed two datasets based on names and summaries, namely \textbf{Movie-Name} and \textbf{Movie-Summary}.
  \item \textbf{GoEmotions\footnote{https://github.com/google-research/google-research/tree/master/goemotions}:}
    It is a corpus of 58k carefully curated comments extracted from Reddit, with human annotations to 27 emotion categories or Neutral.
  \item \textbf{Toxic\footnote{https://www.kaggle.com/competitions/jigsaw-toxic-comment-classification-challenge/data}:}
    This dataset contains a large number of Wikipedia comments which have been labeled by human raters for toxic behavior.
\end{itemize}
We divided the above 11 datasets into training set, validation set and test set according to the ratio of 8:1:1, and the final results are examined on the test set.
See appendix for more information about these datasets.

\subsection{Baselines}\label{s4-2}

We choose the following losses as our comparison objects:
\begin{itemize}
  \item \textbf{The binary cross entropy (BCE)}, which is showed in Equation \ref{eq14}.
  \item \textbf{The focal loss (FL)}:
    \begin{equation}\label{eq19}
      \mathcal{L}_{fl} = -\sum_{i \in \Omega_{pos}}(1-p_i)^{\gamma_{fl}}\log{p_i} -\sum_{j \in \Omega_{neg}}p_j^{\gamma_{fl}}\log(1-p_j), 
    \end{equation}
    where $p_l = \text{sigmoid}(s_l)$.
    We set $\gamma_{fl} = 2$ in experiments.
  \item \textbf{The first version of dice loss (DL1)}:
    \begin{equation}\label{eq20}
      \mathcal{L}_{dl1} = \sum_{i \in \Omega_{pos}}\left(1-\frac{2p_i + \gamma_{dl1}}{p_i^2 + 1 + \gamma_{dl1}}\right) + \sum_{j \in \Omega_{neg}}\left(1-\frac{\gamma_{dl1}}{p_j^2 + \gamma_{dl1}}\right), 
    \end{equation}
    where $p_l = \text{sigmoid}(s_l)$.
    We set $\gamma_{dl1} = 1$ in experiments.
  \item \textbf{The second version of dice loss (DL2)}:
    \begin{equation}\label{eq21}
      \mathcal{L}_{dl2} = \sum_{l = 1}^L\left(1-\frac{2\sum_{k = 1}^Bp_{k, l}y_k^{(l)}+\gamma_{dl2}}{\sum_{k = 1}^Bp_{k, l}^2 + \sum_{k = 1}^By_k^{(l)} + \gamma_{dl2}}\right), 
    \end{equation}
    where $p_{k, l} = \text{sigmoid}(s_{k, l})$ and $k$ stands for the $k$-th sample.
    We set $\gamma_{dl2} = 1$ in experiments.
  \item \textbf{The ranking loss (RL)}:
    \begin{equation}\label{eq22}
      \mathcal{L}_{rl} = \sum_{i \in \Omega_{pos}}\sum_{j \in \Omega_{neg}}\max(0, \alpha_{rl}+s_j-s_i).
    \end{equation}
    We set $\alpha_{rl} = 1$ in experiments.
  \item \textbf{The WARP loss}: 
    \begin{equation}\label{eq23}
      \mathcal{L}_{warp} = \sum_{i \in \Omega_{pos}}\sum_{j \in \Omega_{neg}}w(r_i)\max(0, \alpha_{warp}+s_j-s_i), 
    \end{equation}
    where $w$ is a monotonically increasing function and $r_i$ is the ranking order of $i$-th category.
    We set $w(r_i) = r_i$ and $\alpha_{warp} = 1$ in experiments.
  \item \textbf{The BP-MLL loss}:
    \begin{equation}\label{eq24}
      \mathcal{L}_{bp-mll} = \sum_{i \in \Omega_{pos}}\sum_{j \in \Omega_{neg}}\exp(s_j - s_i).
    \end{equation}
    To ensure numerical stability, we take its logarithmic form, i.e., $\log\mathcal{L}_{bp-mll}$ in experiments.
  \item \textbf{The LSEP loss}, which is showed in Equation \ref{eq12}.
\end{itemize}
The first four are the results of the corresponding binary loss combined with the BR mode, and the last four are ranking-based losses in LR mode.

\subsection{Metrics}\label{s4-3}

We use a variety of evaluation metrics to measure the effect of each loss:
\begin{itemize}
  \item \textbf{The example-based subset accuracy (SubACC)}: 
    \begin{equation}\label{eq25}
      \text{SubACC} = \frac{1}{N}\sum_{n = 1}^NI(\hat{y}_n = y_n), 
    \end{equation}
    where $y_n$ is the $n$-th true label set, $\hat{y}_n$ is the $n$-th predicted label set, and $I$ is the indicative function.
  \item \textbf{The example-based F1-measure (MLC-F1)}:
    \begin{equation}\label{eq26}
      \text{MLC-F1} = \frac{1}{N}\sum_{n = 1}^N\frac{2|\hat{y}_n \cap y_n|}{|\hat{y}_n| + |y_n|}.
    \end{equation}
  \item \textbf{The label-based Macro-F1}:
    \begin{equation}
      \text{Macro-F1} = \frac{1}{L}\sum_{l = 1}^L\frac{2TP_l}{2TP_l+FP_l+FN_l}, 
    \end{equation}
    where $TP_l$ donates the number of True Positive of the $l$-th category, $FP_l$ the number of False Positive, $FN_l$ the number of False Negative.
  \item \textbf{The label-based Micro-F1}:
    \begin{equation}
      \text{Micro-F1} = \frac{2\sum_{l = 1}^LTP_l}{2\sum_{l = 1}^LTP_l + \sum_{l = 1}^LFP_l + \sum_{l = 1}^LFN_l}.
    \end{equation}
  \item \textbf{The ranking-based average precision (AvgPrec)}:
    \begin{equation}
      \text{AvgPrec} = \frac{1}{N}\sum_{n = 1}^N\frac{1}{|y_n|}\sum_{\lambda \in y_n}\frac{|\{\lambda' \in y_n: r_n(\lambda') \leq r_n(\lambda)\}|}{r_n(\lambda)}.
    \end{equation} 
    where $r_n(\lambda)$ is the predicted ranking order of category $\lambda$.
  \item \textbf{The ranking-based ranking loss (RankLoss)}:
    \begin{equation}
      \text{RL} = \frac{1}{N}\sum_{n = 1}^N\frac{1}{|\hat{y}_n||y_n|}|\{(\lambda_a, \lambda_b): r_n(\lambda_a)>r_n(\lambda_b), (\lambda_a, \lambda_b) \in y_n\otimes(\Lambda/y_n)\}|, 
    \end{equation}
\end{itemize}
In the above metrics, the smaller the RL value, the better the result, and the larger the other values, the better the result.

\subsection{Results}

We use Bert-base\cite{DBLP:journals/corr/abs-1810-04805} and Roberta-base\cite{DBLP:journals/corr/abs-1907-11692} as backbones combined with each loss function, and a total of 22 (11 datasets paired with 2 backbones) sets of comparative experiments were carried out.
We evaluate the performance of BCE, FL, DL1, DL2, and ZLPR (ours), which can adapt to the changing number of target categories, on all metrics.
The results are shown in Table \ref{tab1}.
\begin{table}[htbp]
  \centering
  \caption{
    The performance of the 5 loss functions under all metrics.
  }\label{tab1}
  \begin{tabular}{ccccccc}
    \toprule
    ~ & SubACC & MLC-F1 & Micro-F1 & Macro-F1 & AvgPrec & RankLoss \\
    \midrule
    BCE & 0 & 2 & 2 & 2 & 4 & 2 \\
    FL & 0 & 1 & 2 & 6 & 5 & 3 \\
    DL1 & 2 & 0 & 0 & 0 & 0 & 0 \\
    DL2 & 4 & \textbf{10} & \textbf{13} & 6 & 0 & 0 \\
    ZLPR (ours) & \textbf{16} & 9 & 5 & \textbf{8} & \textbf{13} & \textbf{18} \\
    \bottomrule
  \end{tabular}
\end{table}
The numbers in the table are the optimal times for the corresponding loss in 22 comparative experiments.
For example, the ZLPR loss performs very well under the SubACC metric, achieving the best in 16 out of the 22 comparative experiments.
The SubACC metric is very strict on the results and is generally used to measure the model's ability to capture label dependencies\cite{2012OnL}.
The good performance of ZLPR on SubACC demonstrates its ability to capture label dependencies, as described in sub-section \ref{s3-1}.
Obviously, ZLPR has also achieved good performance on the two ranking-based measures, which we can consider is because ZLPR requires the model to rank labels.
On the three metrics of MLC-F1, Micro-F1 and Macro-F1, DL2 is a strong competitor of ZLPR, and even outperforms ZLPR on Micro-F1 by a significant gap.
This may be because DL2 is an approximation of the (negative) F1-measure and optimized using batch data instead of one sample.
Even though, ZLPR still outperforms than BCE, FL and DL1 on those F1-measures.
We also perform tests on four ranking-based losses (RL, WARP, BP-MLL and LSEP) on the three Chinese datasets, of which only the RL value of LSEP slightly outperforms ZLPR on CNIPA-Title, and ZLPR is the best for the rest datasets and metrics.

All experiments are implemented based on Bert4Keras\footnote{https://github.com/bojone/bert4keras}.
Adam\cite{2014Adam} with learning rate $2\times 10^{-4}$ is used as the optimizer and models are trained for 20 epochs.
Details of all experiment results can be found in the appendix section.

\section{Conclusion}\label{s5}

We propose the ZLPR loss to generalize the application of deep learning in multi-label classification and conduct extensive textual experiments.
Compared to some binary losses, the ZLPR loss is able to capture better label dependencies and the ranking relation between positive and negative categories.
Compared with the previous ranking-based losses, ZLPR can adaptively determine the number of target categories, and also strengthen the label ranking ability of models.
However, the performance of ZLPR on F1-measures is not as dazzling as other metrics, which we will improve in the future.

\bibliographystyle{unsrt}  
\bibliography{main}  

\begin{thebibliography}{10}

\bibitem{2017Faster}
Shaoqing Ren, Kaiming He, Ross Girshick, and Jian Sun.
\newblock Faster r-cnn: Towards real-time object detection with region proposal
  networks.
\newblock {\em IEEE Transactions on Pattern Analysis \& Machine Intelligence},
  39(6):1137--1149, 2017.

\bibitem{2020Federated}
Y.~Shi, Y.~Tong, Z.~Su, D.~Jiang, and W.~Zhang.
\newblock Federated topic discovery: A semantic consistent approach.
\newblock {\em Intelligent Systems, IEEE}, PP(99):1--1, 2020.

\bibitem{2021AC}
P.~Eriksson, Nad Marzouka, G.~Sjdahl, C.~Bernardo, F.~Liedberg, and M.~Hglund.
\newblock A comparison of rule-based and centroid single-sample multiclass
  predictors for transcriptomic classification.
\newblock {\em Bioinformatics}, 2021.

\bibitem{Christel2022Multilabel}
Christel~Gérardin a~b, Perceval~Wajsbürt c, Pascal~Vaillant D, Ali~Bellamine
  B, Fabrice Carrat~A E, and Xavier~Tannier C.
\newblock Multilabel classification of medical concepts for patient clinical
  profile identification.
\newblock {\em Artificial Intelligence in Medicine}, 2022.

\bibitem{2017AnB}
Bing Zhu~A B, Bart Baesens~B C, and Seppe~K.L.M. vanden Broucke~b.
\newblock An empirical comparison of techniques for the class imbalance problem
  in churn prediction.
\newblock {\em Information Sciences}, 408:84--99, 2017.

\bibitem{2020Fast}
Y.~Wang, Y.~Xie, Y.~Liu, K.~Zhou, and X.~Li.
\newblock Fast graph convolution network based multi-label image recognition
  via cross-modal fusion.
\newblock In {\em CIKM '20: The 29th ACM International Conference on
  Information and Knowledge Management}, 2020.

\bibitem{2014Towards}
M.~L. Zhang, Y.~K. Li, and X.~Y. Liu.
\newblock Towards class-imbalance aware multi-label learning.
\newblock In {\em the Twenty-Fourth International Joint Conference on
  Artificial Intelligence}, 2014.

\bibitem{bogatinovski2022comprehensive}
Jasmin Bogatinovski, Ljup{\v{c}}o Todorovski, Sa{\v{s}}o D{\v{z}}eroski, and
  Dragi Kocev.
\newblock Comprehensive comparative study of multi-label classification
  methods.
\newblock {\em Expert Systems with Applications}, 203:117215, 2022.

\bibitem{2021Multi}
S.~Li and J.~Ou.
\newblock Multi-label classification of research papers using multi-label
  k-nearest neighbour algorithm.
\newblock {\em Journal of Physics: Conference Series}, 1994(1):012031 (10pp),
  2021.

\bibitem{4634206}
Eneldo Loza~Mencia and Johannes Furnkranz.
\newblock Pairwise learning of multilabel classifications with perceptrons.
\newblock In {\em 2008 IEEE International Joint Conference on Neural Networks
  (IEEE World Congress on Computational Intelligence)}, pages 2899--2906, 2008.

\bibitem{2008Calibrated}
A.~Jiang, C.~Wang, and Y.~Zhu.
\newblock Calibrated rank-svm for multi-label image categorization.
\newblock In {\em IEEE International Joint Conference on Neural Networks},
  2008.

\bibitem{2016DeepBE}
C.~Li, K.~Qi, G.~Ge, S.~Qiang, H.~Lu, and C.~Jian.
\newblock Deepbe: Learning deep binary encoding for multi-label classification.
\newblock In {\em IEEE Conference on Computer Vision \& Pattern Recognition
  Workshops}, 2016.

\bibitem{2015Deep}
F.~Zhao, Y.~Huang, L.~Wang, and T.~Tan.
\newblock Deep semantic ranking based hashing for multi-label image retrieval.
\newblock In {\em 2015 IEEE Conference on Computer Vision and Pattern
  Recognition (CVPR)}, 2015.

\bibitem{2014Deep}
J.~Read and F.~Perez-Cruz.
\newblock Deep learning for multi-label classification.
\newblock {\em Machine Learning}, 85(3):333--359, 2014.

\bibitem{2017Focal}
T.~Y. Lin, P.~Goyal, R.~Girshick, K.~He, and P~Dollár.
\newblock Focal loss for dense object detection.
\newblock {\em IEEE Transactions on Pattern Analysis \& Machine Intelligence},
  PP(99):2999--3007, 2017.

\bibitem{2017Improving}
Y.~Li, Y.~Song, and J.~Luo.
\newblock Improving pairwise ranking for multi-label image classification.
\newblock In {\em 2017 IEEE Conference on Computer Vision and Pattern
  Recognition (CVPR)}, 2017.

\bibitem{2021Delving}
C.~B. Zhang, P.~T. Jiang, Q.~Hou, Y.~Wei, and M.~M. Cheng.
\newblock Delving deep into label smoothing.
\newblock {\em IEEE Transactions on Image Processing}, PP(99):1--1, 2021.

\bibitem{2021arXiv210614448L}
Xiaobo {Liang}, Lijun {Wu}, Juntao {Li}, Yue {Wang}, Qi~{Meng}, Tao {Qin}, Wei
  {Chen}, Min {Zhang}, and Tie-Yan {Liu}.
\newblock {R-Drop: Regularized Dropout for Neural Networks}.
\newblock {\em arXiv e-prints}, page arXiv:2106.14448, June 2021.

\bibitem{2012article}
Mahendra Sahare and Hitesh Gupta.
\newblock A review of multi-class classification for imbalanced data.
\newblock {\em International Journal of Advanced Computer Research},
  2:160--164, 09 2012.

\bibitem{2016V}
F.~Milletari, N.~Navab, and S.~A. Ahmadi.
\newblock V-net: Fully convolutional neural networks for volumetric medical
  image segmentation.
\newblock In {\em 2016 Fourth International Conference on 3D Vision (3DV)},
  2016.

\bibitem{li2019dice}
Xiaoya Li, Xiaofei Sun, Yuxian Meng, Junjun Liang, Fei Wu, and Jiwei Li.
\newblock Dice loss for data-imbalanced nlp tasks.
\newblock {\em arXiv preprint arXiv:1911.02855}, 2019.

\bibitem{menon2020long}
Aditya~Krishna Menon, Sadeep Jayasumana, Ankit~Singh Rawat, Himanshu Jain,
  Andreas Veit, and Sanjiv Kumar.
\newblock Long-tail learning via logit adjustment.
\newblock {\em arXiv preprint arXiv:2007.07314}, 2020.

\bibitem{2016CNN}
W.~Jiang, Y.~Yi, J.~Mao, Z.~Huang, and X.~Wei.
\newblock Cnn-rnn: A unified framework for multi-label image classification.
\newblock In {\em 2016 IEEE Conference on Computer Vision and Pattern
  Recognition (CVPR)}, 2016.

\bibitem{2021History}
A~Yx, L.~A. Yi, Y.~A. Jin, A~Sg, X.~B. Yi, and A~Zl.
\newblock History-based attention in seq2seq model for multi-label text
  classification.
\newblock {\em Knowledge-Based Systems}, 2021.

\bibitem{2020Learning}
H.~Azarbonyad, M.~Dehghani, M.~Marx, and J.~Kamps.
\newblock Learning to rank for multi-label text classification: Combining
  different sources of information.
\newblock {\em Natural Language Engineering}, 27(1):1--23, 2020.

\bibitem{1970Weighted}
L.~J. Hong.
\newblock Weighted approximately ranked pairwise loss (warp).
\newblock {\em Hongliangjie Com}, 1970.

\bibitem{2008Improved}
Rafa Grodzicki, J~Mańdziuk, and L.~Wang.
\newblock Improved multilabel classification with neural networks.
\newblock In {\em International Conference on Parallel Problem Solving from
  Nature: Ppsn X}, 2008.

\bibitem{2020Circle}
Y.~Sun, C.~Cheng, Y.~Zhang, C.~Zhang, L.~Zheng, Z.~Wang, and Y.~Wei.
\newblock Circle loss: A unified perspective of pair similarity optimization.
\newblock In {\em 2020 IEEE/CVF Conference on Computer Vision and Pattern
  Recognition (CVPR)}, 2020.

\bibitem{2020AF}
Eyke Hüllermeier, Marcel Wever, Eneldo~Loza Mencia, Johannes Fürnkranz, and
  Michael Rapp.
\newblock A flexible class of dependence-aware multi-label loss functions,
  2020.

\bibitem{2012OnL}
K.~Dembczyński, W.~Waegeman, and W.~Cheng.
\newblock On label dependence and loss minimization in multi-label
  classification.
\newblock {\em Machine Learning}, 2012.

\bibitem{2013Distributed}
T.~Mikolov, I.~Sutskever, C.~Kai, G.~Corrado, and J.~Dean.
\newblock Distributed representations of words and phrases and their
  compositionality.
\newblock In {\em arXiv}, 2013.

\bibitem{2021GPT}
Robert Dale.
\newblock Gpt-3: What's it good for?
\newblock {\em Natural Language Engineering}, 27(1):113--118, 2021.

\bibitem{DBLP:journals/corr/abs-1810-04805}
Jacob Devlin, Ming{-}Wei Chang, Kenton Lee, and Kristina Toutanova.
\newblock {BERT:} pre-training of deep bidirectional transformers for language
  understanding.
\newblock {\em CoRR}, abs/1810.04805, 2018.

\bibitem{DBLP:journals/corr/abs-1907-11692}
Yinhan Liu, Myle Ott, Naman Goyal, Jingfei Du, Mandar Joshi, Danqi Chen, Omer
  Levy, Mike Lewis, Luke Zettlemoyer, and Veselin Stoyanov.
\newblock Roberta: {A} robustly optimized {BERT} pretraining approach.
\newblock {\em CoRR}, abs/1907.11692, 2019.

\bibitem{2014Adam}
D.~Kingma and J.~Ba.
\newblock Adam: A method for stochastic optimization.
\newblock {\em Computer Science}, 2014.

\bibitem{miyato2018virtual}
Takeru Miyato, Shin-ichi Maeda, Masanori Koyama, and Shin Ishii.
\newblock Virtual adversarial training: a regularization method for supervised
  and semi-supervised learning.
\newblock {\em IEEE transactions on pattern analysis and machine intelligence},
  41(8):1979--1993, 2018.

\end{thebibliography}

\newpage
\begin{appendices}

\section{The Soft-ZLPR Loss}

By simply replacing $y$ with the binary probability vector $p$ of labels, where $p_i$ represents the probability of $i$-th label being positive, we get
\begin{equation}\label{eq31}
  \mathcal{L}_{zlpr}^{soft} = \log\left[1+\langle p, e^{-s}\rangle\right] + \log\left[1+\langle 1-p, e^{s}\rangle\right].
\end{equation}
This is the candidate we need for the soft-label version of the ZLPR loss and what we are going to do next is proving that it works.

In machine learning, an effective loss means that the target of the related task can be rebuild by the output when it gets the optimum.
Taking the derivative with respect to $\mathcal{L}_{zlpr}^{soft}$:
\begin{equation}\label{eq32}
  \frac{\partial\mathcal{L}_{zlpr}^{soft}}{\partial s_i} = -\frac{p_ie^{-s_i}}{1+\langle p, e^{-s}\rangle} + \frac{(1-p_i)e^{s_i}}{1+\langle 1-p, e^{s}\rangle}, 
\end{equation}
and it gets the optimum when
\begin{equation}\label{eq33}
  \forall i: \frac{\partial\mathcal{L}_{zlpr}^{soft}}{\partial s_i} = 0.
\end{equation}
It is tough to solve Equation \ref{eq33} analytically.
Luckily, we find a simple solution by observing:
\begin{equation}\label{eq34}
  p_ie^{-s_i} = (1-p_i)e^{s_i}, 
\end{equation}
which derives
\begin{equation}\label{eq35}
  p_i = \frac{1}{1+e^{-2s_i}} = \text{sigmoid}(2s_i).
\end{equation}

Equation \ref{eq35} is an inspirational result and expresses the following information:
\begin{itemize}
  \item $\mathcal{L}_{zlpr}^{soft}$ is a reasonable extension of the zlpr loss for soft-label, because it can rebuild $p$ by proper $s$.
  \item The transition from scores to binary probabilities should be $\hat{p} = \text{sigmoid}(2s)$ instead of the intuitive $\text{sigmoid}(s)$.
\end{itemize}

\section{The Calculation of KL-divergence}

Some regularization training methods based on sample interference, such as the virtual adversarial training\cite{miyato2018virtual} and R-drop\cite{2021arXiv210614448L}, require the calculation of KL-divergence.
The original output of the ZLPR loss is not a probability distribution but label scores.
However, after the discussion of the soft-label version, we know that label scores can be transformed into the relevant binary probability form.
Equation \ref{eq35} implies that every label is selected or not in MLC can still be treated as a binary classification problem, which means we can use the transformed binary distribution to calculate the KL-divergence.
With the transformation of Equation \ref{eq35}, the KL-divergence between $s$ and $s'$ can be defined as:
\begin{align}\label{eq36}
  \mathcal{KL}(s, s') &\equiv \sum_{i = 1}^n\mathcal{KL}(\hat{p}_i, \hat{p}_i') \\ \nonumber
  &= \sum_{i = 1}^n\left[\hat{p}_i(\log\hat{p}_i-\log\hat{p}_i') + (1-\hat{p}_i)(\log(1-\hat{p}_i)-\log(1-\hat{p}_i'))\right] \\ \nonumber
  &= \sum_{i = 1}^n\left[2\hat{p}_i(s_i-s_i')+\log\frac{1-\hat{p}_i}{1-\hat{p}_i'}\right], 
\end{align}
and the symmetric version is
\begin{equation}\label{eq37}
  \mathcal{D}(s, s') = \mathcal{KL}(s, s') + \mathcal{KL}(s', s) = \sum_{i = 1}^n2(\hat{p}_i-\hat{p}_i')(s_i-s_i').
\end{equation}
Although the form of multiple dichotomies is used to calculate the KL-divergence of the ZLPR loss, it is not affected by the data imbalance caused by the BR mode, because it is not a classification task per se.

\newpage
\section{Statistics of Datasets}

\begin{table}[htbp]
\centering
\caption{
  Statistics of Datasets.
}\label{tab2}
\begin{tabular}{lccccc}
\toprule
\textbf{Dataset} & \textbf{Instances} & \textbf{AveWords} & \textbf{Categories} & \textbf{Cardinality} & \textbf{Density} \\
\midrule
CNIPA-data       & 212, 095      & 17/242       & 618            & 1.330           & 0.002       \\
CNIPA-train      & 169, 626      & 17/242       & 617            & 1.330           & 0.002       \\
CNIPA-val        & 21, 361       & 17/241       & 570            & 1.329           & 0.002       \\
CNIPA-test       & 21, 108       & 17/242       & 572            & 1.331           & 0.002       \\
\midrule
Toutiao-data     & 659, 870      & 20       & 142            & 1.530           & 0.011       \\
Toutiao-train    & 527, 785      & 20       & 142            & 1.530           & 0.011       \\
Toutiao-val      & 65, 861       & 20       & 104            & 1.529           & 0.015       \\
Toutiao-test     & 66, 224       & 20       & 110            & 1.533           & 0.014       \\
\midrule
USPTO-data       & 353, 701      & 8/111        & 632            & 2.174           & 0.003       \\
USPTO-train      & 283, 000      & 8/111        & 628            & 2.175           & 0.003       \\
USPTO-val        & 35, 273       & 8/112        & 601            & 2.171           & 0.004       \\
USPTO-test       & 35, 428       & 8111         & 603            & 2.169           & 0.004       \\
\midrule
Archive-data     & 20, 972       & 10/151       & 6              & 1.252           & 0.209       \\
Archive-train    & 16, 794       & 10/152       & 6              & 1.255           & 0.209       \\
Archive-val      & 2, 045        & 10/151       & 6              & 1.246           & 0.208       \\
Archive-test     & 2, 133        & 10/151       & 6              & 1.239           & 0.207       \\
\midrule
CMU-Movie-data       & 42, 204       & 3/312        & 372            & 3.691           & 0.010       \\
CMU-Movie-train      & 33, 751       & 3/312        & 370            & 3.700           & 0.010       \\
CMU-Movie-val        & 4, 241        & 3/310        & 277            & 3.660           & 0.013       \\
CMU-Movie-test       & 4, 212        & 3/318        & 284            & 3.649           & 0.013       \\
\midrule
GoEmotions-data  & 211, 225      & 13           & 28             & 1.181           & 0.042       \\
GoEmotions-train & 168, 692      & 13           & 28             & 1.181           & 0.042       \\
GoEmotions-val   & 21, 429       & 13           & 28             & 1.183           & 0.042       \\
GoEmotions-test  & 21, 104       & 13           & 28             & 1.182           & 0.042       \\
\midrule
Toxic-data       & 159, 571      & 70           & 6              & 0.220           & 0.037       \\
Toxic-train      & 127, 705      & 70           & 6              & 0.219           & 0.036       \\
Toxic-val        & 15, 855       & 70           & 6              & 0.227           & 0.038       \\
Toxic-test       & 16, 011       & 70           & 6              & 0.224           & 0.037       \\
\bottomrule
\end{tabular}
\end{table}

\newpage
\section{Details of Chinese Text Experiment Results}

\begin{table}[htbp]
\centering
\caption{
  Experiment Results on CNIPA-Title.
}\label{tab3}
\begin{tabular}{lcccccc}
\toprule
\multicolumn{7}{c}{Bert}                                             \\
\midrule
Loss  & SubACC  & MLC-F1  & Micro-F1 & Macro-F1 & AvgPrec & RankLoss \\
\midrule
BCE   & 42.75\% & 56.09\% & 57.72\%  & 39.79\%  & 70.69\% & 1.12\%   \\
FL    & 41.71\% & 56.10\% & 57.76\%  & \textbf{44.21\%}  & 71.06\% & 1.21\%   \\
DL1   & 36.28\% & 46.48\% & 52.06\%  & 21.86\%  & 59.76\% & 6.72\%   \\
DL2   & 41.83\% & 55.08\% & 57.56\%  & 36.42\%  & 68.05\% & 2.66\%   \\
RL    &         &         &          &          & 71.59\% & 1.24\%   \\
WARP  &         &         &          &          & 71.59\% & 1.32\%   \\
BP-MLL &         &         &          &          & 70.24\% & 1.07\%   \\
LSEP  &         &         &          &          & 69.29\% & \textbf{1.01\%}   \\
ZLPR  & \textbf{43.23\%} & \textbf{57.54\%} & \textbf{58.34\%}  & 43.47\%  & \textbf{71.68\%} & 1.02\%   \\
\midrule
\multicolumn{7}{c}{Roberta}                                          \\
\midrule
Loss  & SubACC  & MLC-F1  & Micro-F1 & Macro-F1 & AvgPrec & RankLoss \\
\midrule
BCE   & 42.41\% & 56.12\% & 57.70\%  & 40.83\%  & 70.41\% & 1.17\%   \\
FL    & 41.18\% & 55.92\% & 57.59\%  & 44.22\%  & 70.92\% & 1.16\%   \\
DL1   & 37.58\% & 48.28\% & 53.49\%  & 23.51\%  & 62.16\% & 5.85\%   \\
DL2   & 42.47\% & 56.03\% & \textbf{58.23\%}  & 40.06\%  & 68.66\% & 2.74\%   \\
RL    &         &         &          &          & 71.33\% & 1.51\%   \\
WARP  &         &         &          &          & 71.32\% & 1.46\%   \\
BP-MLL &         &         &          &          & 70.10\% & 1.05\%   \\
LSEP  &         &         &          &          & 69.80\% & \textbf{0.99\%}   \\
ZLPR  & \textbf{42.79\%} & \textbf{57.66\%} & 58.20\%  & \textbf{44.17\%}  & \textbf{71.45\%} & 1.00\%  \\
\bottomrule
\end{tabular}
\end{table}

\begin{table}[htbp]
\centering
\caption{
  Experiment Results on CNIPA-Abstract.
}\label{tab4}
\begin{tabular}{lcccccc}
\toprule
\multicolumn{7}{c}{Bert}                                             \\
\midrule
Loss  & SubACC  & MLC-F1  & Micro-F1 & Macro-F1 & AvgPrec & RankLoss \\
\midrule
BCE     & 45.23\% & 59.81\% & 61.01\%  & 40.02\%  & 74.02\% & 0.86\%   \\
FL      & 45.26\% & 60.78\% & 61.75\%  & \textbf{46.09\%}  & 75.08\% & 0.88\%   \\
DL1     & 40.18\% & 52.01\% & 57.00\%  & 23.47\%  & 64.48\% & 6.53\%   \\
DL2     & 46.65\% & 61.77\% & \textbf{62.95\%}  & 42.98\%  & 73.42\% & 2.24\%   \\
RL      &         &         &          &          & 76.27\% & 0.88\%   \\
WARP    &         &         &          &          & 76.35\% & 0.84\%   \\
BP-MLL   &         &         &          &          & 74.10\% & 0.85\%   \\
LSEP    &         &         &          &          & 74.05\% & 0.70\%   \\
ZLPR    & \textbf{47.25\%} & \textbf{62.62\%} & 62.68\%  & 45.75\%  & \textbf{76.41\%} & \textbf{0.64\%}   \\
\midrule
\multicolumn{7}{c}{Roberta}                                          \\
\midrule
Loss  & SubACC  & MLC-F1  & Micro-F1 & Macro-F1 & AvgPrec & RankLoss \\
\midrule
BCE     & 45.17\% & 58.99\% & 60.69\%  & 38.69\%  & 73.85\% & 0.88\%   \\
FL      & 44.65\% & 59.54\% & 61.09\%  & 44.85\%  & 74.36\% & 0.98\%   \\
DL1     & 41.56\% & 53.55\% & 58.03\%  & 25.34\%  & 65.84\% & 5.65\%   \\
DL2     & 46.66\% & 61.30\% & \textbf{62.72\%}  & 42.57\%  & 73.13\% & 2.13\%   \\
RL      &         &         &          &          & 76.18\% & 0.91\%   \\
WARP    &         &         &          &          & 76.18\% & 1.06\%   \\
BP-MLL   &         &         &          &          & 74.15\% & 0.83\%   \\
LSEP    &         &         &          &          & 74.64\% & 0.71\%   \\
ZLPR    & \textbf{46.86\%} & \textbf{62.47\%} & 62.64\%  & \textbf{47.12\%}  & \textbf{76.24\%} & \textbf{0.68\%}  \\
\bottomrule
\end{tabular}
\end{table}

\begin{table}[htbp]
\centering
\caption{
  Experiment Results on Toutiao.
}\label{tab5}
\begin{tabular}{lcccccc}
\toprule
\multicolumn{7}{c}{Bert}                                             \\
\midrule
Loss  & SubACC  & MLC-F1  & Micro-F1 & Macro-F1 & AvgPrec & RankLoss \\
\midrule
BCE     & 47.76\% & 68.40\% & 69.74\%  & 43.14\%  & 82.35\% & 0.89\%   \\
FL      & 47.40\% & 67.88\% & 69.52\%  & 43.34\%  & 82.32\% & 0.87\%   \\
DL1     & 48.13\% & 67.86\% & 69.40\%  & 42.78\%  & 81.32\% & 1.16\%   \\
DL2     & 47.64\% & 69.28\% & 69.35\%  & 42.97\%  & 81.18\% & 1.11\%   \\
BP-MLL   &         &         &          &          & 81.62\% & 0.92\%   \\
LSEP    &         &         &          &          & 81.19\% & 0.93\%   \\
RL      &         &         &          &          & 81.81\% & 1.06\%   \\
WARP    &         &         &          &          & 81.69\% & 0.91\%   \\
ZLPR    & \textbf{48.29\%} & \textbf{69.49\%} & \textbf{69.85\%}  & \textbf{43.84\%}  & \textbf{82.52\%} & \textbf{0.85\%}   \\
\midrule
\multicolumn{7}{c}{Roberta}                                          \\
\midrule
Loss  & SubACC  & MLC-F1  & Micro-F1 & Macro-F1 & AvgPrec & RankLoss \\
\midrule
BCE     & 47.77\% & 68.68\% & 69.91\%  & 43.45\%  & 82.49\% & 0.87\%   \\
FL      & 47.46\% & 68.08\% & 69.53\%  & \textbf{43.90\%}  & 82.29\% & 0.88\%   \\
DL1     & 48.01\% & 67.53\% & 69.32\%  & 42.21\%  & 81.68\% & 1.13\%   \\
DL2     & 47.71\% & \textbf{69.51\%} & 69.52\%  & 42.89\%  & 81.35\% & 1.10\%   \\
BPMLL   &         &         &          &          & 81.76\% & 0.88\%   \\
LSEP    &         &         &          &          & 81.19\% & 0.92\%   \\
RL      &         &         &          &          & 81.94\% & 1.00\%   \\
WARP    &         &         &          &          & 81.97\% & 0.92\%   \\
ZLPR    & \textbf{48.40\%} & 69.19\% & \textbf{69.93\%}  & 43.79\%  & \textbf{82.77\%} & \textbf{0.83\%}   \\
\bottomrule
\end{tabular}
\end{table}

\newpage
\section{Details of English Text Experiment Results}

\begin{table}[htbp]
\centering
\caption{
  Experiment Results on USPTO-Title.
}\label{tab6}
\begin{tabular}{lcccccc}
\toprule
\multicolumn{7}{c}{Bert}                                             \\
\midrule
Loss  & SubACC  & MLC-F1  & Micro-F1 & Macro-F1 & AvgPrec & RankLoss \\
\midrule
BCE     & 23.54\% & 50.06\% & 50.17\%  & 25.48\%  & 65.26\% & 2.03\%   \\
FL      & 23.34\% & 50.54\% & 50.76\%  & \textbf{29.77\%}  & \textbf{66.08\%} & 1.98\%   \\
DL1     & 23.61\% & 48.15\% & 49.27\%  & 19.21\%  & 61.53\% & 4.83\%   \\
DL2     & 24.10\% & \textbf{51.84\%} & \textbf{52.06\%}  & 29.08\%  & 63.60\% & 3.58\%   \\
ZLPR    & \textbf{24.34\%} & 50.97\% & 50.00\%  & 26.92\%  & 66.05\% & \textbf{1.87\%}   \\
\midrule
\multicolumn{7}{c}{Roberta}                                          \\
\midrule
Loss  & SubACC  & MLC-F1  & Micro-F1 & Macro-F1 & AvgPrec & RankLoss \\
\midrule
BCE     & 27.57\% & \textbf{57.88\%} & 55.72\%  & 34.19\%  & 71.53\% & 1.37\%   \\
FL      & 27.28\% & 57.26\% & 55.57\%  & \textbf{35.01\%}  & \textbf{71.84\%} & 1.31\%   \\
DL1     & 27.51\% & 54.37\% & 53.49\%  & 23.95\%  & 68.21\% & 3.42\%   \\
DL2     & 27.30\% & 57.62\% & \textbf{56.10\%}  & 31.50\%  & 69.62\% & 2.40\%   \\
ZLPR    & \textbf{27.81\%} & 56.40\% & 53.93\%  & 30.85\%  & 71.32\% & \textbf{1.26\%}   \\
\bottomrule
\end{tabular}
\end{table}

\begin{table}[htbp]
\centering
\caption{
  Experiment Results on USPTO-Abstract.
}\label{tab7}
\begin{tabular}{lcccccc}
\toprule
\multicolumn{7}{c}{Bert}                                             \\
\midrule
Loss  & SubACC  & MLC-F1  & Micro-F1 & Macro-F1 & AvgPrec & RankLoss \\
\midrule
BCE     & 31.44\% & 63.94\% & 60.98\%  & 34.13\%  & 77.31\% & 0.84\%   \\
FL      & 31.20\% & 64.27\% & 61.43\%  & \textbf{38.03\%}  & 77.91\% & 0.84\%   \\
DL1     & 31.37\% & 62.42\% & 60.50\%  & 26.23\%  & 72.65\% & 3.65\%   \\
DL2     & 31.93\% & \textbf{65.06\%} & \textbf{62.23\%}  & 35.02\%  & 75.85\% & 1.85\%   \\
ZLPR    & \textbf{32.57\%} & 64.93\% & 61.42\%  & 36.56\%  & \textbf{78.67\%} & \textbf{0.72\%}   \\
\midrule
\multicolumn{7}{c}{Roberta}                                          \\
\midrule
Loss  & SubACC  & MLC-F1  & Micro-F1 & Macro-F1 & AvgPrec & RankLoss \\
\midrule
BCE     & 31.72\% & 64.73\% & 61.70\%  & 36.36\%  & 78.29\% & 0.79\%   \\
FL      & 31.72\% & 64.32\% & 61.48\%  & 38.27\%  & 78.49\% & 0.75\%   \\
DL1     & 31.72\% & 61.98\% & 59.77\%  & 26.57\%  & 74.16\% & 2.95\%   \\
DL2     & 31.81\% & \textbf{65.54\%} & \textbf{62.53\%}  & 38.18\%  & 76.33\% & 1.72\%   \\
ZLPR    & \textbf{32.61\%} & 64.84\% & 61.48\%  & \textbf{38.60\%}  & \textbf{78.71\%} & \textbf{0.71\%}   \\
\bottomrule
\end{tabular}
\end{table}

\begin{table}[htbp]
\centering
\caption{
  Experiment Results on Archive-Title.
}\label{tab8}
\begin{tabular}{lcccccc}
\toprule
\multicolumn{7}{c}{Bert}                                             \\
\midrule
Loss  & SubACC  & MLC-F1  & Micro-F1 & Macro-F1 & AvgPrec & RankLoss \\
\midrule
BCE     & 49.79\% & 62.59\% & 64.26\%  & 52.14\%  & 81.58\% & 11.84\%  \\
FL      & 48.57\% & 62.74\% & 64.62\%  & 52.60\%  & \textbf{81.61\%} & \textbf{11.68\%}  \\
DL1     & 48.62\% & 60.88\% & 63.47\%  & 53.35\%  & 81.00\% & 12.81\%  \\
DL2     & 47.77\% & \textbf{66.77\%} & \textbf{66.59\%}  & \textbf{56.20\%}  & 81.50\% & 11.89\%  \\
ZLPR    & \textbf{50.02\%} & 64.98\% & 65.42\%  & 54.03\%  & 81.50\% & 12.03\%  \\
\midrule
\multicolumn{7}{c}{Roberta}                                          \\
\midrule
Loss  & SubACC  & MLC-F1  & Micro-F1 & Macro-F1 & AvgPrec & RankLoss \\
\midrule
BCE     & 59.59\% & 74.73\% & 75.22\%  & 68.41\%  & 87.16\% & 7.61\%   \\
FL      & 59.21\% & \textbf{75.83\%} & \textbf{76.05\%}  & 68.55\%  & \textbf{87.64\%} & 7.34\%   \\
DL1     & \textbf{60.99\%} & 74.39\% & 75.09\%  & 67.51\%  & 87.28\% & 7.73\%   \\
DL2     & 58.23\% & 75.12\% & 74.59\%  & \textbf{69.09\%}  & 86.60\% & 8.09\%   \\
ZLPR    & 60.38\% & 75.76\% & 75.73\%  & 68.93\%  & 87.48\% & \textbf{7.32\%}   \\
\bottomrule
\end{tabular}
\end{table}

\begin{table}[htbp]
\centering
\caption{
  Experiment Results on Archive-Abstract.
}\label{tab9}
\begin{tabular}{lcccccc}
\toprule
\multicolumn{7}{c}{Bert}                                             \\
\midrule
Loss  & SubACC  & MLC-F1  & Micro-F1 & Macro-F1 & AvgPrec & RankLoss \\
\midrule
BCE     & 67.28\% & 81.87\% & 80.98\%  & 73.85\%  & 91.28\% & 4.90\%   \\
FL      & 66.81\% & 81.89\% & \textbf{81.42\%}  & \textbf{74.49\%}  & \textbf{91.61\%} & 4.56\%   \\
DL1     & 67.28\% & 81.66\% & 80.85\%  & 71.93\%  & 91.19\% & 5.55\%   \\
DL2     & 66.81\% & \textbf{82.29\%} & 80.96\%  & 74.24\%  & 91.11\% & 5.36\%   \\
ZLPR    & \textbf{67.37\%} & 82.27\% & 81.23\%  & 73.40\%  & 91.45\% & \textbf{4.74\%}   \\
\midrule
\multicolumn{7}{c}{Roberta}                                          \\
\midrule
Loss  & SubACC  & MLC-F1  & Micro-F1 & Macro-F1 & AvgPrec & RankLoss \\
\midrule
BCE     & 67.46\% & 82.53\% & 81.72\%  & 76.42\%  & \textbf{91.73\%} & \textbf{4.53\%}   \\
FL      & 65.07\% & 81.39\% & 81.07\%  & 71.12\%  & 91.54\% & 4.70\%   \\
DL1     & 65.17\% & 81.40\% & 80.89\%  & 72.86\%  & 91.31\% & 4.87\%   \\
DL2     & 66.85\% & 82.37\% & 81.55\%  & \textbf{76.74\%}  & 91.43\% & 4.96\%   \\
ZLPR    & \textbf{67.75\%} & \textbf{82.59\%} & \textbf{81.63\%}  & 74.26\%  & 91.68\% & 4.63\%   \\
\bottomrule
\end{tabular}
\end{table}

\begin{table}[htbp]
\centering
\caption{
  Experiment Results on Movie-Name.
}\label{tab10}
\begin{tabular}{lcccccc}
\toprule
\multicolumn{7}{c}{Bert}                                             \\
\midrule
Loss  & SubACC  & MLC-F1  & Micro-F1 & Macro-F1 & AvgPrec & RankLoss \\
\midrule
BCE     & 2.80\% & 7.85\%  & 7.25\%   & 23.84\%  & \textbf{37.48\%} & 5.25\%   \\
FL      & 2.83\% & 8.25\%  & 7.83\%   & 23.93\%  & 37.06\% & 5.27\%   \\
DL1     & 1.47\% & 2.30\%  & 1.26\%   & 23.77\%  & 37.34\% & 5.55\%   \\
DL2     & \textbf{6.03\%} & \textbf{21.80\%} & \textbf{19.27\%}  & 23.88\%  & 36.29\% & 5.97\%   \\
ZLPR    & 5.53\% & 20.65\% & 18.97\%  & \textbf{24.19\%}  & 37.04\% & \textbf{5.18\%}   \\
\midrule
\multicolumn{7}{c}{Roberta}                                          \\
\midrule
Loss  & SubACC  & MLC-F1  & Micro-F1 & Macro-F1 & AvgPrec & RankLoss \\
\midrule
BCE     & 5.39\% & \textbf{26.74\%} & \textbf{27.57\%}  & 25.85\%  & \textbf{44.47\%} & 4.34\%   \\
FL      & 5.82\% & 23.63\% & 23.40\%  & 25.10\%  & 43.95\% & 4.29\%   \\
DL1     & 5.25\% & 21.92\% & 22.72\%  & 24.99\%  & 43.03\% & 5.16\%   \\
DL2     & \textbf{6.03\%} & 21.80\% & 19.27\%  & 23.88\%  & 36.31\% & 6.06\%   \\
ZLPR    & 5.51\% & 26.09\% & 26.55\%  & \textbf{25.93\%}  & 44.08\% & \textbf{4.17\%}   \\
\bottomrule
\end{tabular}
\end{table}

\begin{table}[htbp]
\centering
\caption{
  Experiment Results on Movie-Summary.
}\label{tab11}
\begin{tabular}{lcccccc}
\toprule
\multicolumn{7}{c}{Bert}                                             \\
\midrule
Loss  & SubACC  & MLC-F1  & Micro-F1 & Macro-F1 & AvgPrec & RankLoss \\
\midrule
BCE     & 7.24\% & 32.06\% & 33.42\%  & 26.34\%  & 53.70\% & 3.28\%   \\
FL      & 7.24\% & 35.06\% & 36.46\%  & 27.65\%  & 56.03\% & 2.59\%   \\
DL1     & 6.98\% & 34.29\% & 35.02\%  & 26.22\%  & 51.25\% & 4.95\%   \\
DL2     & 5.98\% & 40.26\% & \textbf{41.34\%}  & 28.16\%  & 52.96\% & 3.74\%   \\
ZLPR    & \textbf{7.43\%} & \textbf{40.34\%} & 40.55\%  & \textbf{28.57\%}  & \textbf{56.55\%} & \textbf{2.49\%}   \\
\midrule
\multicolumn{7}{c}{Roberta}                                          \\
\midrule
Loss  & SubACC  & MLC-F1  & Micro-F1 & Macro-F1 & AvgPrec & RankLoss \\
\midrule
BCE     & 8.24\% & 39.02\% & 40.98\%  & 28.40\%  & 58.27\% & 2.36\%   \\
FL      & 7.53\% & 40.10\% & 41.96\%  & 29.97\%  & 58.86\% & 2.10\%   \\
DL1     & 7.76\% & 36.07\% & 38.24\%  & 27.23\%  & 56.46\% & 3.96\%   \\
DL2     & 7.55\% & \textbf{43.88\%} & \textbf{45.48\%}  & 30.80\%  & 56.94\% & 3.00\%   \\
ZLPR    & \textbf{8.50\%} & 42.04\% & 43.20\%  & \textbf{31.32\%}  & \textbf{59.98\%} & \textbf{1.98\%}   \\
\bottomrule
\end{tabular}
\end{table}

\begin{table}[htbp]
\centering
\caption{
  Experiment Results on GoEmotions.
}\label{tab12}
\begin{tabular}{lcccccc}
\toprule
\multicolumn{7}{c}{Bert}                                             \\
\midrule
Loss  & SubACC  & MLC-F1  & Micro-F1 & Macro-F1 & AvgPrec & RankLoss \\
\midrule
BCE     & 24.96\% & 31.18\% & 35.83\%  & 29.77\%  & 53.86\% & 12.08\%  \\
FL      & 23.92\% & 32.14\% & 34.47\%  & 29.09\%  & 51.73\% & 13.36\%  \\
DL1     & 25.87\% & 30.56\% & 35.36\%  & 26.00\%  & 53.65\% & 13.67\%  \\
DL2     & \textbf{29.08\%} & \textbf{38.25\%} & \textbf{40.49\%}  & 30.42\%  & 56.28\% & 11.67\%  \\
ZLPR    & 27.01\% & 33.07\% & 37.72\%  & \textbf{31.29\%}  & \textbf{56.68\%} & \textbf{10.23\%}  \\
\midrule
\multicolumn{7}{c}{Roberta}                                          \\
\midrule
Loss  & SubACC  & MLC-F1  & Micro-F1 & Macro-F1 & AvgPrec & RankLoss \\
\midrule
BCE     & 25.50\% & 32.15\% & 36.44\%  & 31.04\%  & 54.36\% & 11.60\%  \\
FL      & 23.74\% & 32.71\% & 35.30\%  & 29.85\%  & 52.22\% & 13.19\%  \\
DL1     & 26.68\% & 31.56\% & 37.45\%  & 27.93\%  & 56.43\% & 11.66\%  \\
DL2     & \textbf{31.01\%} & \textbf{41.45\%} & \textbf{43.79\%}  & \textbf{33.42\%}  & 59.51\% & 9.83\%   \\
ZLPR    & 29.41\% & 35.22\% & 40.44\%  & 31.54\%  & \textbf{60.00\%} & \textbf{8.89\%}   \\
\bottomrule
\end{tabular}
\end{table}

\begin{table}[htbp]
\centering
\caption{
  Experiment Results on Toxic.
}\label{tab13}
\begin{tabular}{lcccccc}
\toprule
\multicolumn{7}{c}{Bert}                                             \\
\midrule
Loss  & SubACC  & MLC-F1  & Micro-F1 & Macro-F1 & AvgPrec & RankLoss \\
\midrule
BCE     & 90.69\% & 93.86\% & 69.91\%  & \textbf{57.27\%}  & \textbf{9.79\%}  & \textbf{0.35\%}   \\
FL      & 91.06\% & 94.02\% & 69.51\%  & 56.50\%  & 9.63\%  & 0.47\%   \\
DL1     & \textbf{91.75\%} & 94.30\% & 69.30\%  & 54.63\%  & 9.55\%  & 0.51\%   \\
DL2     & 91.34\% & 94.14\% & 68.87\%  & 54.50\%  & 9.59\%  & 0.48\%   \\
ZLPR    & 91.51\% & \textbf{94.38\%} & \textbf{70.38\%}  & 50.62\%  & 9.78\%  & 0.36\%   \\
\midrule
\multicolumn{7}{c}{Roberta}                                          \\
\midrule
Loss  & SubACC  & MLC-F1  & Micro-F1 & Macro-F1 & AvgPrec & RankLoss \\
\midrule
BCE     & 91.96\% & 95.13\% & 76.76\%  & \textbf{65.92\%}  & 9.82\%  & 0.31\%   \\
FL      & 91.86\% & 95.12\% & 76.61\%  & 60.95\%  & 9.83\%  & \textbf{0.30\%}   \\
DL1     & 92.16\% & 95.13\% & 75.47\%  & 54.85\%  & 9.70\%  & 0.45\%   \\
DL2     & 92.05\% & 95.15\% & 76.27\%  & 63.36\%  & 9.82\%  & 0.31\%   \\
ZLPR    & \textbf{92.17\%} & \textbf{95.24\%} & \textbf{77.36\%}  & 64.33\%  & \textbf{9.84\%}  & \textbf{0.30\%}   \\
\bottomrule
\end{tabular}
\end{table}

\end{appendices}

\end{document}